%% file: output.tex
\documentclass[letterpaper, 10 pt, journal, twoside]{IEEEtran}

\usepackage{graphics} 
\usepackage{epsfig} 
\usepackage{amsmath} 
\usepackage{bm}
\usepackage{amssymb}  
\usepackage{algorithm, algorithmic}
\usepackage{xcolor}
\usepackage{graphicx,booktabs,array}
\usepackage[hidelinks]{hyperref}
\usepackage{caption}
\usepackage{subcaption}

\usepackage{physics}

\input{symbols.tex}    

\title{Distributing Collaborative Multi-Robot Planning with Gaussian Belief Propagation}

\begin{document}

\markboth{IEEE Robotics and Automation Letters. Preprint Version. Accepted November, 2022}
{Patwardhan \MakeLowercase{\textit{et al.}}: Distributing Collaborative Multi-Robot Planning with Gaussian Belief Propagation} 

\author{Aalok Patwardhan$^{1}$, Riku Murai$^{2}$ and Andrew J. Davison$^{1}$
\thanks{Manuscript received: August 27, 2022; Revised November 5, 2022; Accepted November 29, 2022.}
\thanks{This paper was recommended for publication by Editor M. Ani Hsieh upon evaluation of the Associate Editor and Reviewers' comments. This work was supported by Dyson Technology Ltd.} 
\thanks{$^{1}$Aalok Patwardhan and Andrew J. Davison are with the Dyson Robotics Lab and the Department of Computing, Imperial College London
        {\tt\footnotesize [a.patwardhan21,a.davison]@imperial.ac.uk}}%
\thanks{$^{2} $Riku Murai is with the Department of Computing, Imperial College London
        {\tt\footnotesize riku.murai15@imperial.ac.uk}}%
\thanks{Digital Object Identifier (DOI): 10.1109/LRA.2022.3227858}
}

\maketitle

\begin{abstract}

Precise coordinated planning over a forward time window enables safe and highly efficient motion when many robots must work together in tight spaces, but this would normally require centralised control of all devices which is difficult to scale. We demonstrate GBP Planning, a new purely distributed technique based on Gaussian Belief Propagation for multi-robot planning problems, formulated by a generic factor graph defining dynamics and collision constraints over a forward time window. In simulations, we show that our method allows high performance collaborative planning where robots are able to cross each other in busy, intricate  scenarios. They maintain shorter, quicker and smoother trajectories than alternative distributed planning techniques even in cases of communication failure. We encourage the reader to view the accompanying video demonstration at \href{https://youtu.be/8VSrEUjH610}{https://youtu.be/8VSrEUjH610}.

\end{abstract}

\section{INTRODUCTION}
\IEEEPARstart{A}{s} automation increases, multiple robotic devices will need to operate together, whether working robots in a home, farm or industrial setting, or autonomous vehicles on a road network. 
To achieve both safety and efficiency when speeds are high and space is limited, these robots must {\em coordinate} when planning their motions and other actions. At the most basic level, robots should take minimal account of each other's plans in order to avoid collisions. When coordination is stronger, extremely high efficiency is possible, and large numbers of devices can move smoothly and intricately around each other as they carry out their tasks.

It might be assumed that highly coordinated planning requires centralised control, where a single computer receives state information from all robots and solves a unified multiple trajectory optimisation problem before sending commands back to them all. In this paper we show that precise performance is in fact possible with an alternative distributed technique which uses efficient local per-robot computation and purely peer-to-peer communication.

Our solution formulates multi-robot planning as a general dynamic optimisation problem characterised by a factor graph with variables representing robot positions and velocities in Euclidean space over a bounded forward time window. These variables are connected by constraint factors which represent each robot's dynamics as well as the need to avoid collisions with other robots and static obstacles. 
Collaborative planning consists of performing inference on the factor graph to determine marginal distributions for the future poses which best satisfy all constraints.

\begin{figure}[ht!]
    \centering
    \includegraphics[width=0.9\linewidth]{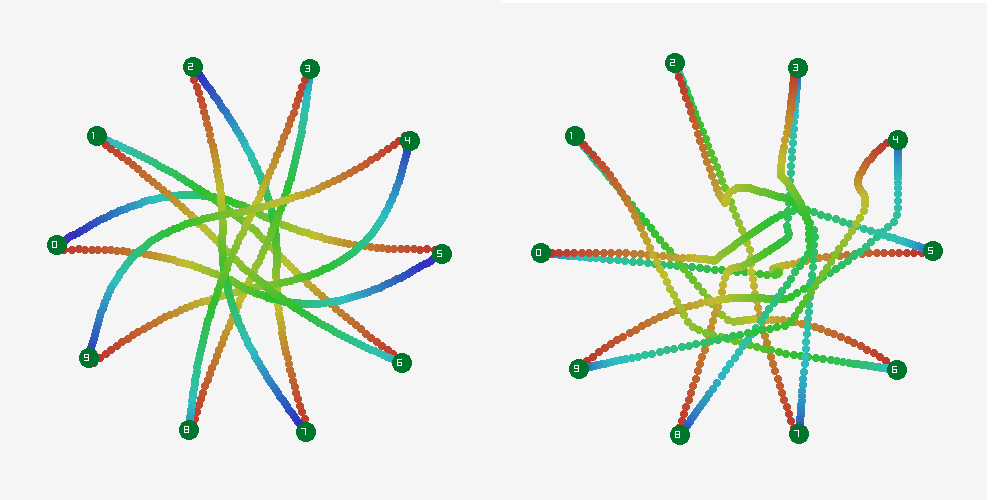}
    \caption{Experiment where 10 robots cross to opposite sides of a circle without colliding. Our GBP planner (left) achieves shorter and smoother trajectories than ORCA (right). Colours change along the paths with time, from red to blue.}
    \label{fig:paths_N10}%
\end{figure}

The key novelty of our approach is to show that the inference required for planning within this general factor graph framework can be achieved via distributed computation. 
We divide the full joint factor graph into fragments which individual robots have responsibility for storing and updating. Inference is then achieved using Gaussian Belief Propagation (GBP), an iterative and distributed message passing algorithm which can achieve the same global solution as a centralised solver. Messages which need to be sent between the graph fragments held by different robots are transmitted by regular peer-to-peer communication of small messages. The particularly appealing property of GBP for multi-robot systems is that global convergence can be achieved with an ad-hoc communication schedule, as would often be the case in real scenarios when for instance robots only achieve intermittent radio contact.

Our GBP Planning algorithm is generic, but we demonstrate it here in simulations of two scenarios of multi-robot planning in tight spaces.  In the first, robots must  simultaneously exchange places across a circle formation to move from one fixed configuration to another. In the  second, two continuous streams of robots meet at a busy junction and aim to cross each other while maintaining high flow rates. In both, the robots jointly plan and update their trajectories online to achieve smooth (dynamically realistic) and efficient trajectories, and we quantitatively measure performance significantly beyond that demonstrated by alternative distributed planning techniques. We also show that planning performance only degrades slowly 
when peer-to-peer communication is reduced or unreliable with a  large fraction of messages lost.

In summary, our key contributions are:
\bi
\item The first use of Gaussian Belief Propagation as a general method for distributed multi-robot planning with multiple dynamics and obstacle constraints over a forward window. This leads to a simple and highly generic technique.
\item Quantitative and qualitative simulation experiments of 2D many-robot scenarios that show that the intricate multi-robot coordination achieved by GBP Planning significantly outperforms a well known alternative distributed planning method in this setting.
\ei

\section{RELATED WORK}


In  single-robot planning, a robot solves  for a path towards  its goal while balancing the need to move in accordance with  dynamics limits and to avoid obstacles. These various aspects are expressed as hard or soft constraints in a cost function, and an algorithm then solves for the trajectory with the minimum total cost. Is often computationally intractable to find the global minimum cost trajectory for complex problems, but iterative solvers can find locally optimal solutions for a finite number of look-ahead steps. 

Multi-robot planning is a generalisation of this, where simultaneous trajectory solutions are sought for multiple robots.
Multi-robot planning in tight spaces entails conflict resolution because each robot's future planned positions become obstacles for every other robot. Centralised approaches (e.g. 
\cite{Soria:etal:NMI2021})
 can achieve high performance, but require all robots to be coordinated by a single hub; this represents clear challenges to scalability. 
We therefore focus here on distributed methods like ours, which can operate with per-robot parallelism and peer-to-peer communication.

One possible approach is priority-based planning such as in \cite{vandeberg_priority}. Here, high priority robots largely use single-robot planning methods. Lower-priority robots then treat them as dynamic obstacles with unchanging trajectories which must be avoided, and will therefore often have to wait before being able to reach their goals.
While this makes sense in some applications, there are many other situations
 where there is no meaningful priority ordering.
 Here conflict resolution becomes more subtle because every change to one robot's plan could affect all the others, and the rest of the methods we discuss here apply to this case.

A widely used distributed multi-robot planner is the Optimal Reciprocal Collision Avoidance (ORCA) algorithm \cite{ORCA}. In ORCA, every robot has perfect instantaneous knowledge of the position and velocities of its neighbours. Robots react accordingly by changing their own velocities, but cannot make decisions to resolve conflicts which may arise further into the future. ORCA is fast and has been shown to be effective in some real applications. However, as we will show in comparative results later, ORCA produces trajectories which are often jerky and inefficient when many robots crowd into tight spaces.
A related method where all robots treat each other as velocity obstacles but considering more general kinodynamic robot motions was presented in
\cite{AlonsoMora:etal:TRO2018}. The distributed version of this method only allowed a short look ahead and was demonstrated with very small numbers of robots.
Another interesting distributed method, but also without general communication for resolution of multi-robot forward plans, was presented in
\cite{Senbaslar:etal:DARS2019}
and was demonstrated on six real differential drive robots.

Several existing methods have used iterative communication to enable longer look-ahead.
The work in \cite{vanparys} presents an online solver using Alternating Direction Method of Multipliers (ADMM) for a small number of robots moving through an environment with moving obstacles. Here the robots move together in formation and their paths do not intersect so few conflicts need to be resolved.
The distributed model predictive control algorithm in \cite{Luis:etal:RAL2020}
has been demonstrated for impressive criss-crossing behaviour in teams of real quadrotor robots, though again the gaps between robots are generally larger than we are able to achieve with our approach.
The graph neural network method in \cite{prorok} tackles learned multi-robot planning with local communication, though so far only in a very abstract grid-world setting.



In our work we show for the first time a way to achieve a solution to distributed planning using a general multi-robot cost function over an arbitrary length forward time window.
We are inspired in particular by methods like GPMP2 (the Gaussian Process Motion Planner)
\cite{Mukadam:etal:IJRR2018}. This is a planner for a single robot in a known static environment, but importantly showed that planning with dynamics
can commonly be formulated using a cost function with a sum of quadratic terms, and therefore that solving for a trajectory is equivalent to performing
inference on a Gaussian factor graph. This means that general and powerful factor graph solvers such as GTSAM (Georgia Tech Smoothing and Mapping)~\cite{Dellaert:AR2021} which are normally used for estimation could be easily adapted to planning applications. The definition of all cost terms in a planning cost function as quadratic in a set of variables means that planning is solved by the same sparse least-squares inference computation as arises in robot estimation problems such as  SLAM.

Very recently, Murai \etal ~showed in their Robot Web work \cite{Murai:etal:ARXIV2022} that a highly distributed solution to the distributed estimation problem of multi-robot localisation is possible using efficient distributed computation via Gaussian Belief Propagation and peer-to-peer message passing. 
In our work, we adapt this method for the first time to multi-robot {\em planning}, and show that it is possible to take advantage of the same distributed computation structure. We will show that this enables highly coordinated multi-robot motion planning performance while remaining very flexible and general.

We would like to make clear that expressing a multi-robot planning cost function as a sum of quadratic terms is a highly general formulation. The relative strengths of the terms can all be set as required such that some constraints are much more necessary to obey than others (especially collision avoidance). The quadratic terms can also include non-linear functions in variable space. While this formulation does not strictly allow `hard' constraints, it allows arbitrarily strong ones, and this is true of other distributed planning algorithms such as 
\cite{Luis:etal:RAL2020}
when used in practice.

\section{THEORETICAL BACKGROUND}

\begin{figure}
  \centering
 \includegraphics[width=0.9\linewidth]{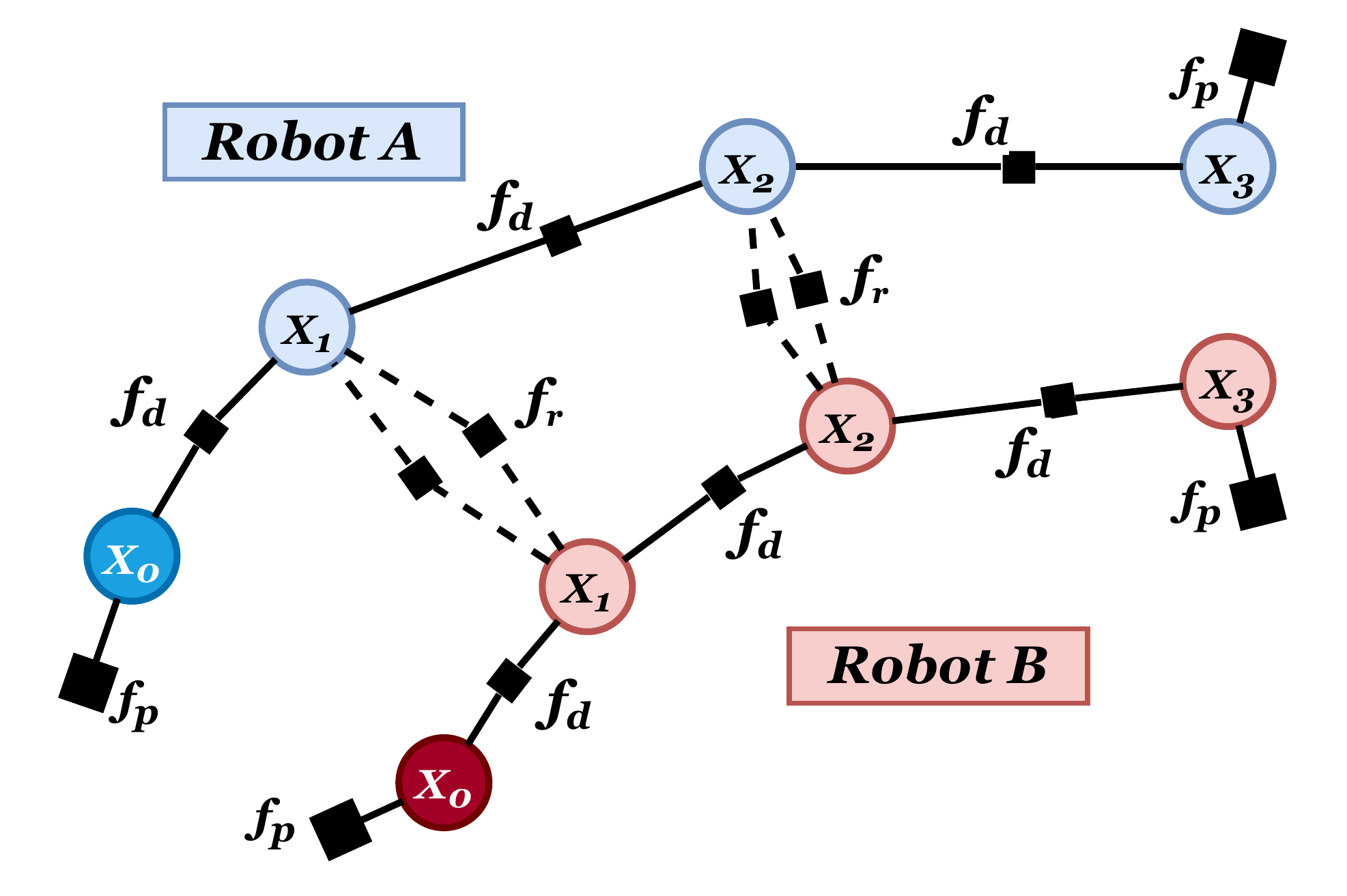}
  \caption{Multi-robot planning factor graph with two robots (A and B) moving from bottom-left to top-right highlighting pose ($f_p$), dynamics ($f_d$) and inter-robot ($f_r$) factors. Unary static obstacle factors $f_o$ are connected to each robot's states but for clarity are not shown here.}
  \label{fig:factorgraph}
\end{figure}

\subsection{Factor Graphs}

A factor graph represents the factorisation of a joint function $p(\bm{\mathrm{X}})$  into components $f_s$ which depend on subsets $\bm{\mathrm{X}}_s$ of all variables $\bm{\mathrm{X}}$: 
\beq
\label{eqn:factorprod}
p(\bm{\mathrm{X}}) = \prod_s f_s(\bm{\mathrm{X}}_s)
~.
\eeq
Graphically, it is an undirected bipartite graph, where each factor is represented by a square node attached by edges to circular nodes representing the variables it involves.

In a Gaussian factor graph all factors have the form of Gaussian distributions:
\beq
\label{equ:generalfactor}
f_s(\bm{\mathrm{X}}_s) = K e^{-\frac{1}{2} \left[ (\bm{z}_s - \bm{h}_s(\bm{\mathrm{X}}_s))^\top \bm{\Lambda}_s (\bm{z}_s - \bm{h}_s(\bm{\mathrm{X}}_s))   \right]  }
~,
\eeq
where $\bm{h}_s(\bm{\mathrm{X}}_s)$ is the (possibly non-linear) functional  form of the measurement or constraint that the factor represents, $\bm{z}_s$ is its observed or expected value, and $\bm{\Lambda}_s$ is the  precision (inverse  covariance) of the constraint and $K$ is a constant.

If we take the negative log of $p(\bm{\mathrm{X}})$ to form the `energy' $E(\bm{\mathrm{X}}) = -\log p(\bm{\mathrm{X}})$, we  obtain:
\bea
E(\bm{\mathrm{X}}) = L + \frac{1}{2} \sum_s   (\bm{z}_s - \bm{h}_s(\bm{\mathrm{X}}_s))^\top \bm{\Lambda}_s (\bm{z}_s - \bm{h}_s(\bm{\mathrm{X}}_s))
\label{eqn:factorsum}
\eea
where $L$ is a constant.

Factor graph inference, for instance in probabilistic estimation problems such as  SLAM, involves finding the values of variables $\bm{\mathrm{X}}$ which  maximise the product  $p(\bm{\mathrm{X}})$. Clearly this is equivalent to finding $\bm{\mathrm{X}}$ to {\em  minimise} $E(\bm{\mathrm{X}}) = -\log p(\bm{\mathrm{X}})$. Equation \ref{eqn:factorsum} shows that a for a  Gaussian factor graph, this  comes down to minimising a sum of squared terms.

Whenever a planning problem can be formulated as a least squares minimisation, it is equivalent to performing inference on a Gaussian Factor Graph.
This opens up the  use  for planning of the many factor graph inference tools which were  originally developed for  probabilistic inference. This  was highlighted in   
\cite{Mukadam:etal:IJRR2018,Dellaert:AR2021}, but previously only demonstrated for single robot planning using centralised solvers such as GTSAM \cite{Dellaert:AR2021}. In this work we use a factor graph to model multi-robot planning, as illustrated in figure~\ref{fig:factorgraph}. We will explain the details of the factors shown in Section~\ref{sec:method}.

\subsection{Gaussian Belief Propagation (GBP)}

GBP allows inference on Gaussian factor graphs via distributed computation. Convergence is achieved via node-wise computation and message passing around the graph, so storage of the graph can be shared between multiple devices communicating by radio or some other channel. 
GBP is the special case of more general loopy belief propagation and empirically has excellent performance, obtaining exact solutions for the marginal means of all variables with rapid convergence~\cite{Ortiz:etal:ARXIV2021}.
In multi-robot systems, the recent Robot Web approach \cite{Murai:etal:ARXIV2022} showed how GBP could be used for
decentralised and scalable localisation of groups of up to 1000 robots making noisy inter-robot measurements.
In this paper, we now show for the first time that thanks to the common mathematical formulation of factor graphs, GBP can alternatively be used for multi-robot {\em planning}.

GBP proceeds by iterating message passing between variables and factors around the loopy factor graph. At any time the current marginal beliefs for variables can be obtained. Here we lay out the main steps; 
see~\cite{Murai:etal:ARXIV2022}, \cite{Ortiz:etal:ARXIV2021} for full details.

We represent a Gaussian distribution in canonical/information form as:
\beq
    \mathcal{N}(\bm{\mathrm{X}}; \bm{\mu}, \bm{\Sigma}) = \mathcal{N}^{-1}(\bm{\mathrm{X}}; \bm{\eta}, \bm{\Lambda})~,
\eeq
where  $\bm{\Lambda} = \bm{\Sigma}^{-1}$ and $\bm{\eta} = \bm{\Lambda} \bm{\mu}$ are the precision matrix and information vector respectively.
In GBP, variables $\bm{\mathrm{X}}_s$ are assumed to be Gaussian; thus, each variable $\bm{\mathrm{x}}_k$ has a belief  $b(\bm{\mathrm{x}}_k) = \mathcal{N}^{-1}(\bm{\mathrm{x}}_k; \bm{\eta}_k, \bm{\Lambda}_k)$. 
Factors $F = \{f_s\}_{s=1:N_f}$ are Gaussian constraints between variables; the factor $f_s(\bm{\mathrm{X}}_s)$ is an arbitrary function that connects variables $\bm{\mathrm{X}}_s$, and it may be non-linear.

\subsubsection{Variable Belief Update}
\label{sec:variable_belief_update}
A variable $\bm{\mathrm{x}}_k$ updates its belief by taking the product of all incoming messages from its connected factors:
\beq
b(\bm{\mathrm{x}}_k) = \prod_{f\in n(\bm{\mathrm{x}}_k)} \vecm_{f\rightarrow k}(\bm{\mathrm{x}}_k)~,
\eeq
where $n(\bm{\mathrm{x}}_k) \subseteq F$ is the set of factors that the variable $\bm{\mathrm{x}}_k$ is connected to, and $\bm{\mathrm{m}}_{f\rightarrow k}(\bm{\mathrm{x}}_k) = \mathcal{N}^{-1}(\bm{\mathrm{x}}_k; \bm{\eta}_{f\rightarrow k}, \bm{\Lambda}_{f\rightarrow k})$ is the message from a factor to the variable. In the canonical or information representation of Gaussians, this product can be rewritten as a summation:
\bea 
\bm{\eta}_{k} &=& \sum_{f\in n(\bm{\mathrm{x}}_k)} \bm{\eta}_{f\rightarrow k}~, \\
\bm{\Lambda}_{k} &=& \sum_{f\in n(\bm{\mathrm{x}}_k)} \bm{\Lambda}_{f\rightarrow k}~.
\eea

\subsubsection{Variable to Factor Message}
A message from a variable $\bm{\mathrm{x}}_k$ to a factor $f_j \in n(\bm{\mathrm{x}}_k)$ is the product of all incoming factor to variable messages apart from the message from $f_j$:
\beq
\vecm_{\bm{\mathrm{x}}_k\rightarrow j}(f_j) = \prod_{f\in n(\bm{\mathrm{x}}_k) \backslash f_j} \vecm_{f\rightarrow k}(\bm{\mathrm{x}}_k)~.
\eeq

\subsubsection{Factor Likelihood Update}
The likelihood of a factor $f_s(\bm{\mathrm{X}}_s)$ connected to variables $\bm{\mathrm{X}}_s$ with  measurement function $\bm{h}_s(\bm{\mathrm{X}}_s)$, observation $\bm{z}_s$, and precision of the observation $\bm{\Lambda}_s$, can be expressed as a Gaussian distribution $\mathcal{N}^{-1}(\bm{\mathrm{X}}_s; \bm{\eta}_f, \bm{\Lambda}_f)$, where $\bm{\eta}_f = \bm{\Lambda}_s(\bm{\mathrm{z}}_s - \bm{h}_s(\bm{\mathrm{X}}_s))$ and $\bm{\Lambda}_f = \bm{\Lambda}_s$.
This however only holds if $\bm{h}_s(\bm{\mathrm{X}}_s)$ is linear. In the non-linear case, we linearise using first-order Taylor expansion: $\bm{h}_s(\bm{\mathrm{X}}_s) \approx \bm{h}_s(\bm{\mathrm{X}}_s^0) + \bm{J}_s(\bm{\mathrm{X}}_s - \bm{\mathrm{X}}_s^0)$. The likelihood of the linearised factor takes the form~\cite{Davison:Ortiz:ARXIV2019}:
\bea
    \bm{\eta}_f &=& \bm{J}_s^{\top}\bm{\Lambda}_s\left( \bm{J}_s \bm{\mathrm{X}}_s^0 + \bm{\mathrm{z}}_s - \bm{h}_s(\bm{\mathrm{X}}_s^0)   \right)~, \\
    \bm{\Lambda}_f &=& \bm{J}_s^{\top}\bm{\Lambda}_s \bm{J}_s~,
\eea
where $\bm{J}_s$ is the Jacobian and $\bm{\mathrm{X}}_s^0$ is the linearisation point: the current state of the variables.
In our work $\bm{\mathrm{z}}_s=0$ for all the factors unless stated otherwise, meaning that the factor energy is purely a function of the states. 

\subsubsection{Factor to Variable Message}
A message from a factor to variable $\bm{\mathrm{x}}_k$ is:
\beq
\vecm_{f\rightarrow k}(\bm{\mathrm{x}}_k) = \sum_{\bm{\mathrm{x}} \in \bm{\mathrm{X}}_s \backslash \bm{\mathrm{x}}_k} f_s(\bm{\mathrm{X}}_s) \prod_{\bm{\mathrm{x}} \in \bm{\mathrm{X}}_s \backslash \bm{\mathrm{x}}_k} \vecm_{\bm{\mathrm{x}}\rightarrow f}(\bm{\mathrm{x}})~.
\eeq
We take the product of the factor likelihood and messages from $\bm{\mathrm{X}}_s \backslash \bm{\mathrm{x}}_k$, then marginalise out all variables but $\bm{\mathrm{x}}_k$.

\section{METHOD}
\label{sec:method}

We now explain the details of how we formulate multi-robot planning as a factor graph. Although GBP can optimise factor graphs for any type of Gaussian variables and factors, for the rest of the paper we will consider a 2D scenario where robots are modelled as objects with two degrees of movement freedom on a plane.

The state $\bm{\mathrm{x}}_k$ of a robot at time $t_k$ represents the robot’s position and velocity at that particular moment in time:
\begin{equation}
\bm{\mathrm{x}}_k=[\bm{x}_k^\top,\bm{\dot{x}}_k^\top]^\top=[x_k,y_k,\dot{x}_k,\dot{y}_k]^\top~.
\end{equation}
The short-term planned trajectory \bm{\mathrm{X}} of a robot with a lookahead horizon of $t_{K-1}$ seconds is parameterised by $K$ such states at discrete times $t_k$ into the future, from the current state $\bm{\mathrm{x}}_0$ of the robot at the current time $t_0$ to the horizon state $\bm{\mathrm{x}}_{K-1}$ at time $t_{K-1}$. Each state is a variable in a factor graph and is connected to the previous state in time with a factor representing dynamics. The states are therefore all Markovian in time, resulting in a sparse linear system.
The optimal solution for the planned trajectory (the poses from the current state until the lookahead horizon) can be found by solving for the maximum a posteriori (MAP) solution $\bm{\mathrm{X}^*}$ over the trajectory states. Our method also determines marginal covariances for these planned states. A high covariance will arise in wide open areas where many solutions have similar cost, and a small one where precise movement is needed to satisfy constraints. 

In the presence of moving obstacles and other robots the safety of the robot (collision avoidance) is paramount. It is the states in the immediate future that are of more importance when optimising a trajectory. The time gaps between consecutive states therefore increase for states that are further along the planned trajectory.

There are four types of factors that we consider. A unary pose factor $f_{p}$ is connected to the current and horizon states of the robot and represents the prior information on those states. A dynamics factor $f_{d}$ encourages a smooth trajectory and represents a noise-on-acceleration dynamics model. An obstacle factor $f_{o}$ penalises a state from being close to a stationary obstacle in the scene. Finally, an inter-robot factor $f_{r}$ penalises trajectories that bring two robots within collision range at a particular timestep, and takes into account each robot’s future positions. An example of the resulting factor graph may be seen in figure \ref{fig:factorgraph}. 


\subsection{Pose Factor}
We define a unary pose factor $f_p$ connected to the current and horizon states of the robot with measurement function and precision matrix:
\begin{gather}
    \bm{h}_p(\bm{\mathrm{x}}_k) = \bm{\mathrm{x}}_k~,    \\
    \bm{\Lambda}_p = \sigma_p^{-2} \bm{\mathrm{I}} ~.
\end{gather}
The observation $\bm{z}_k$ is a constant set to the desired anchoring value of the state $\bm{\mathrm{x}}^{0}_k$.
The robot current and horizon states ($\bm{\mathrm{x}}_0$ and $\bm{\mathrm{x}}_{K-1}$) are connected to strong pose factors with small values of $\sigma_p$, ensuring that the trajectory is anchored at these states during optimisation at a timestep. At the next timestep, these pose factors are modified to reflect the fact that the robot has moved (changed its current state).

\subsection{Dynamics Factor}
\label{ref:dynamics_factor}
For a trajectory to be smooth and dynamically feasible, each state of the robot is connected to the next in time via a factor $f_d$ defined in \cite{Mukadam:etal:IJRR2018} as:
\begin{gather}
    \bm{h}_d(\bm{\mathrm{x}}_k, \bm{\mathrm{x}}_{k+1}) = \bm{\Phi}(t_{k+1}, t_k)\bm{\mathrm{x}}_k - \bm{\mathrm{x}}_{k+1}~,    \\
    \bm{\Lambda}_d = \begin{bmatrix}
        \frac{1}{3}\Delta t_{k}^3 \bm{\mathrm{Q}}_d & \frac{1}{2}\Delta t_{k}^2 \bm{\mathrm{Q}}_d \\
        \frac{1}{2}\Delta t_{k}^2 \bm{\mathrm{Q}}_d & \Delta t_{k} \bm{\mathrm{Q}}_d \end{bmatrix}^{-1},
\end{gather}
with $\Delta t_{k} = t_{k+1} - t_{k}$~, $\bm{\mathrm{Q}}_d = \sigma_d^2 \bm{\mathrm{I}}$~,
and:
\begin{equation}
    \bm{\Phi}(t_b, t_a) = \begin{bmatrix}
        \bm{\mathrm{I}} & (t_b - t_a)\bm{\mathrm{I}} \\
        \bm{0} & \bm{\mathrm{I}} \end{bmatrix}~,
\end{equation}
is the state transition matrix from time $t_a$ to time $t_b$. This factor is derived from a noise-on-acceleration model of dynamics, encouraging a zero acceleration (and therefore a feasible and smooth) trajectory.

\subsection{Obstacle Factor}
A unary factor $f_o$ is connected to each robot state representing its distance from static obstacles in the environment \cite{Mukadam:etal:IJRR2018}. A pre-computed 2D signed distance field (SDF) of the environment is used. The obstacle factor is defined as:
\begin{gather}
    \label{eqn:ho}
    \bm{h}_o\left(\bm{\mathrm{x}}_{k}\right) = \begin{cases}1 - \frac{d_o(\bm{x}_k)}{r_R} & d_o(\bm{x}_k) \leq r_R \\
                        0 & \text{otherwise}\end{cases} , \\
    \bm{\Lambda}_o = \sigma_o^{-2}\bm{\mathrm{I}}.
\end{gather}
where $d_o(\bm{x}_k)$ is the signed distance of point $\bm{x}_k$ from the nearest static obstacle and $r_R$ is the robot radius. It is assumed that a robot can obtain the signed distance instantaneously from the SDF.

\subsection{Inter-robot Factor}
\label{sec:interrobot_factor}
If a robot A encounters another robot B within its communication range $r_C$, inter-robot factors $f_r$ are created for all states  $\bm{\mathrm{x}}_{k}$ of robot A's planned trajectory and their counterparts in that of robot B for $1\leq k < {K-1}$. This ensures that the two robots will plan to not occupy the same position at the same time (collide).

This factor has non-zero energy if the distance between the robots at a particular timestep is less than a critical distance $r^{*} = 2r_{R} + \epsilon$, where $r_{R}$ is the radius of a robot and $\epsilon$ is a small `safety distance'. The inter-robot factor is defined as:
\begin{equation}
    \label{eqn:hr}
    \bm{h}_r\left(\bm{\mathrm{x}}_{k}^{A}, \bm{\mathrm{x}}_{k}^{B}\right) = \begin{cases}1 - \frac{d_r(\bm{x}_k^A,\bm{x}_k^B)}{r^*} & d_r(\bm{x}_k^A,\bm{x}_k^B) \leq r^* \\
                        0 & \text{otherwise}\end{cases}
\end{equation}
where:
\begin{equation}
    d_r(\bm{x}_k^A,\bm{x}_k^B) = \left\|\bm{x}_{k}^{A} - \bm{x}_{k}^{B}\right\| 
\end{equation}
is the distance between the two robots at timestep $t_{k}$. We augment the notation for robot states with the superscripts $A$ and $B$ to distinguish between the two robots.
The factor precision matrix takes the form 
$\bm{\Lambda}_r = (t_k\sigma_{r})^{-2}\bm{\mathrm{I}}$, such that the factor is weaker for states that are further in the future.

Figure \ref{fig:factorgraph} shows that there are two inter-robot factors for each timestep of the trajectories. This symmetry in factor creation represents a shared responsibility between the two robots for planning safe trajectories but is also a redundancy in the design. In future work this could be extended further to allow heterogeneous robots to define their own trajectory parameters, such as safety thresholds or self-importance.

\subsection{Online Algorithm}
 Each robot performs Algorithm \ref{algo}.
 At each timestep, the current state $\bm{\mathrm{x}}_0$ of each robot is 
 updated using the simulation timestep duration $\Delta T$.
 In the case of a moving horizon, the horizon state $\bm{\mathrm{x}}_{K-1}$ is updated similarly.
 
 When new robots are within communication range $r_C$, inter-robot factors are created between them if they do not already exist, and destroyed when the robots are out of range.
 
 Finally robots perform iterative trajectory optimisation using GBP under two regimes of message passing:
 \subsubsection{Internal GBP}
 $M_I$ iterations of message passing are conducted between the states of a robot, and their associated pose, dynamics and obstacle factors (any factors that are not connected to states of other robots).
 We imagine that an individual robot can be continuously performing internal message passing, and so $M_I$ can be arbitrarily large.

 \subsubsection{Inter-robot GBP}
 $M_R$ iterations of message passing are conducted between the states of one robot that are connected to the states of another robot via the inter-robot factors $f_r$.

 The bottleneck in a distributed multi-robot system is likely to be inter-robot communication, so we would want the number of iterations $M_R$ of inter-robot GBP to be low. This number represents the amount of negotiation between robots.
 
\begin{algorithm}
\caption{For one robot $R_i$}\label{algo}
\begin{algorithmic}[1]
 \STATE {Update $\bm{\mathrm{x}}_0$ and $\bm{\mathrm{x}}_{K-1}$ by $\Delta T$}. \\
 \STATE \textit{Let $C(R_i)$ be a set of robots currently connected to $R_i$.}\\
 \STATE \textit{Let $N(R_i) = \{R_j\ |\ \norm{R_i - R_j} < r_C  \}$ be a set of robots within the communication radius of $R_i$.}\\
 \WHILE{Running}
 \FOR {$R_j \in N(R_i) \backslash C(R_i)$}
 \STATE {Create inter-robot factors $f_r(R_i, R_j)$ with a newly observed robot $R_j$.}
 \ENDFOR
 \FOR {$R_j \in C(R_i) \backslash N(R_i)$}
 \STATE {Delete inter-robot factors $f_r(R_i, R_j)$ with a robot $R_j$ out of range.}
 \ENDFOR
 \STATE Perform $M_I$ internal and $M_R$ inter-robot GBP iterations to adapt trajectory to the newly updated states and information.
 \ENDWHILE
\end{algorithmic} 
\end{algorithm}

\section{EVALUATION}
\label{sec:evaluation}
Our method for multi-robot planning is generalised and can be applied to many scenarios in which robots traverse a 2D environment. We consider scenarios where coordination between robots is of importance. 

In the first experiment robots of various sizes begin in a circle formation and travel to the antipodal sides. We also investigate the effect of obstacles and the communication range of robots. In the second experiment we investigate the performance of our algorithm for robots traversing a junction, typical of highly coordinated robots in a warehouse space. 
Finally we highlight the robustness of GBP planning to communications failure due to dropped messages. 

\subsection{Baseline}
We compare our method with the popular Optimal Reciprocal Collision Avoidance (ORCA) \cite{ORCA} which is a multi-agent distributed collision avoidance system that can be used in problems requiring a high degree of coordination.
With ORCA each robot uses Linear Programming (LP) to optimise its instantaneous velocity based on the constraints imposed by the velocities of its neighbours. If no solution can be found ORCA chooses the velocity that violates the least number of constraints.

Our method is also distributed and can react to new information as the robot follows its trajectory but unlike ORCA is a short term state planner. In our experiments we consider robots moving in tight spaces at speeds an order of magnitude higher than those used in the ORCA demonstrations. As such there were three main parameters in the ORCA algorithm that we tuned for our experiments to allow for a fair and meaningful comparison to our method.
\subsubsection{neighbourDist}
The maximum distance to other robots that a robot takes into account when optimizing for a new safe velocity. The larger this number the greater number of constraints on the robot. For a fair comparison of ORCA with our method we set this parameter to be equal to the communication range $r_C$.

\subsubsection{timeHorizon} the smallest time into the future for which collision-free paths are guaranteed. We set this to the value of the horizon $T_{K-1}$ used in our method.

\subsubsection{timeStep} if the simulation timestep was too large, collisions occurred owing to the high speeds of the robots and so a value of $\Delta T=0.05$~s was used for ORCA.

\subsection{Metrics}
We compare metrics for efficiency and smoothness of trajectories. Robots working together in high-coordination problems must minimise their distance travelled as well as the total time taken to reach their goals. The planned trajectories (which are constantly being replanned) must also be smooth and collision-free. As a robot moves through the environment we calculate the following metrics:

\subsubsection{Distance travelled}
The total distance travelled by the robot until its goal is reached. Efficient trajectories should minimise this metric. 
\subsubsection{Makespan}
The total time taken for all robots to reach their goals. A system of many robots working together efficiently should minimise this metric.
\subsubsection{Smoothness}
Robot trajectories must be smooth, in order to be feasible with respect to real-world constraints such as those due to actuators. Since smoothness is intrinsically a geometric property of the path, it should not depend on time taken or velocity. We use the Log Dimensionless Jerk (LDJ), a smoothness metric first defined in \cite{LDJ_paper}. This metric is a log-normalised value of the total squared jerk along the trajectory and is defined as
\begin{equation}
    LDJ~\overset{\Delta}{=}~-\ln\left(\frac{(t_{final} - t_{start})^3}{v_{max}^2} \int_{t_{start}}^{t_{final}}\left|\ddot{v}(t)\right|^2dt\right)
\end{equation}
where $t \in [t_{start}, t_{final}]$ is the time interval over which the metric is considered, $v(t)$ is the robot velocity at time $t$, and $v_{max}$ is the maximum velocity along the trajectory.
The LDJ metric was shown to be a better indicator of smoothness than other dimensionless metrics, and values that are more positive represent `smoother' paths.

\section{Results and Discussion}
For the following experiments we set the simulation timestep to be $\Delta T=0.1$~s for our GBP planner. Inter-robot and internal  GBP was conducted with $M_R=10$ and $M_I=50$ iterations per timestep respectively. We also set $\sigma_p=1\times10^{-15}$~m and $\sigma_r=\sigma_o=0.005$. We encourage the reader to view the accompanying video demonstration.

\subsection{Circle Experiment}
Robots of various sizes are initialised in a circle formation of radius $50$~m, with an initial speed of $15$~m/s, towards a stationary horizon state at the opposite side of the circle. The radii of the robots are sampled randomly from $r_R\sim\mathcal{U}(2,3)$~m.
We set $t_{K-1}$ as the ideal time taken for a single robot moving from the initial speed to rest across the circle at constant acceleration. This would correspond to a smooth (zero jerk) trajectory.
Each robot has a communication range $r_C=50$~m representing a partially connected network of robots. In addition, we set $\sigma_d=1$~m.

This is a complex problem requiring coordination among many robots. Figure \ref{fig:paths_N10} shows the paths taken by GBP planning and ORCA when $N_R=10$. GBP planning allows robots to collaborate and reach their goals at similar times.

\begin{figure}
    \centering
    \includegraphics[width=0.8\linewidth]{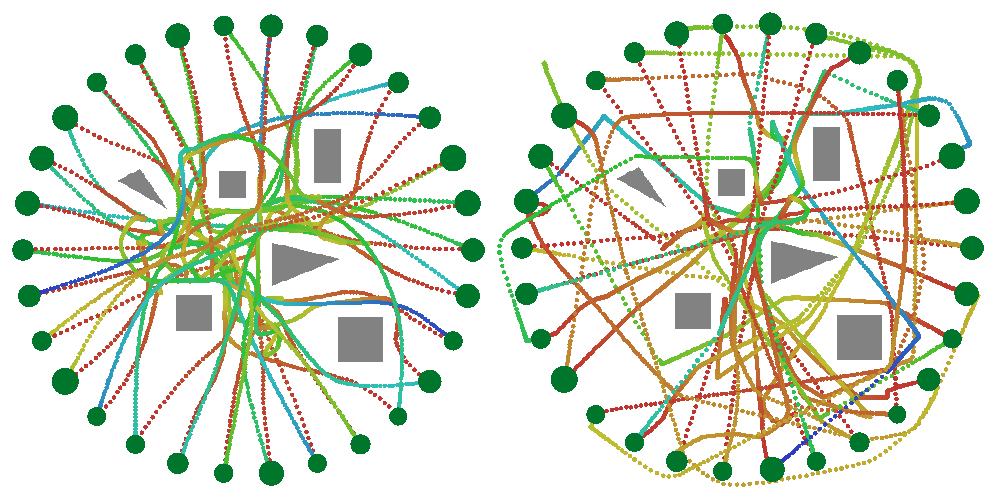}
    \caption{Circle experiment with obstacles for $N_R=30$ robots for the GBP planner (left) and ORCA (right). Colours change along the paths with time, from red (oldest) to blue (newest).}
    \label{fig:paths_N30}%
\end{figure}

\begin{figure}
    \centering
    \includegraphics[width=0.85\linewidth]{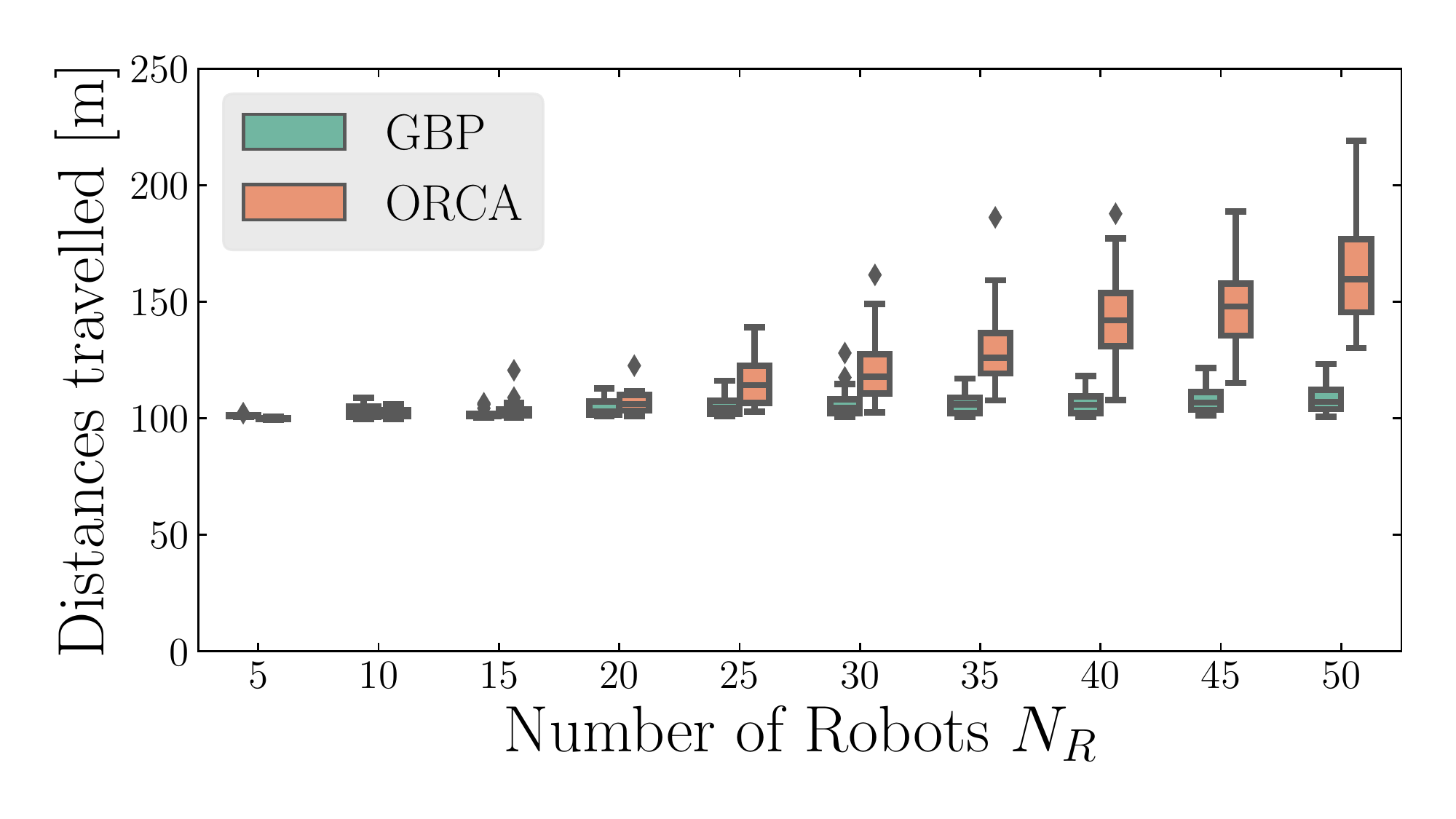}
    \caption{
        Circle experiment: distribution of distances travelled as number of robots in the formation $N_R$ is varied. The GBP planner creates shorter paths and a smaller spread of distances than ORCA; robots collaborate to achieve their goals.
    }
    \label{fig:distances}
\end{figure}

As the number of robots $N_R$ is varied, figure \ref{fig:distances} shows the distribution of robot distances travelled. With the GBP planner, path lengths were close to $100$~m (the diameter of the circle) and were shorter than with ORCA. The spread of robot distances travelled was also smaller due to inter-robot collaboration, whereas in ORCA each robot acted for itself, choosing a path at the expense of longer paths for others.

\begin{figure}
    \centering
    \includegraphics[width=0.85\linewidth]{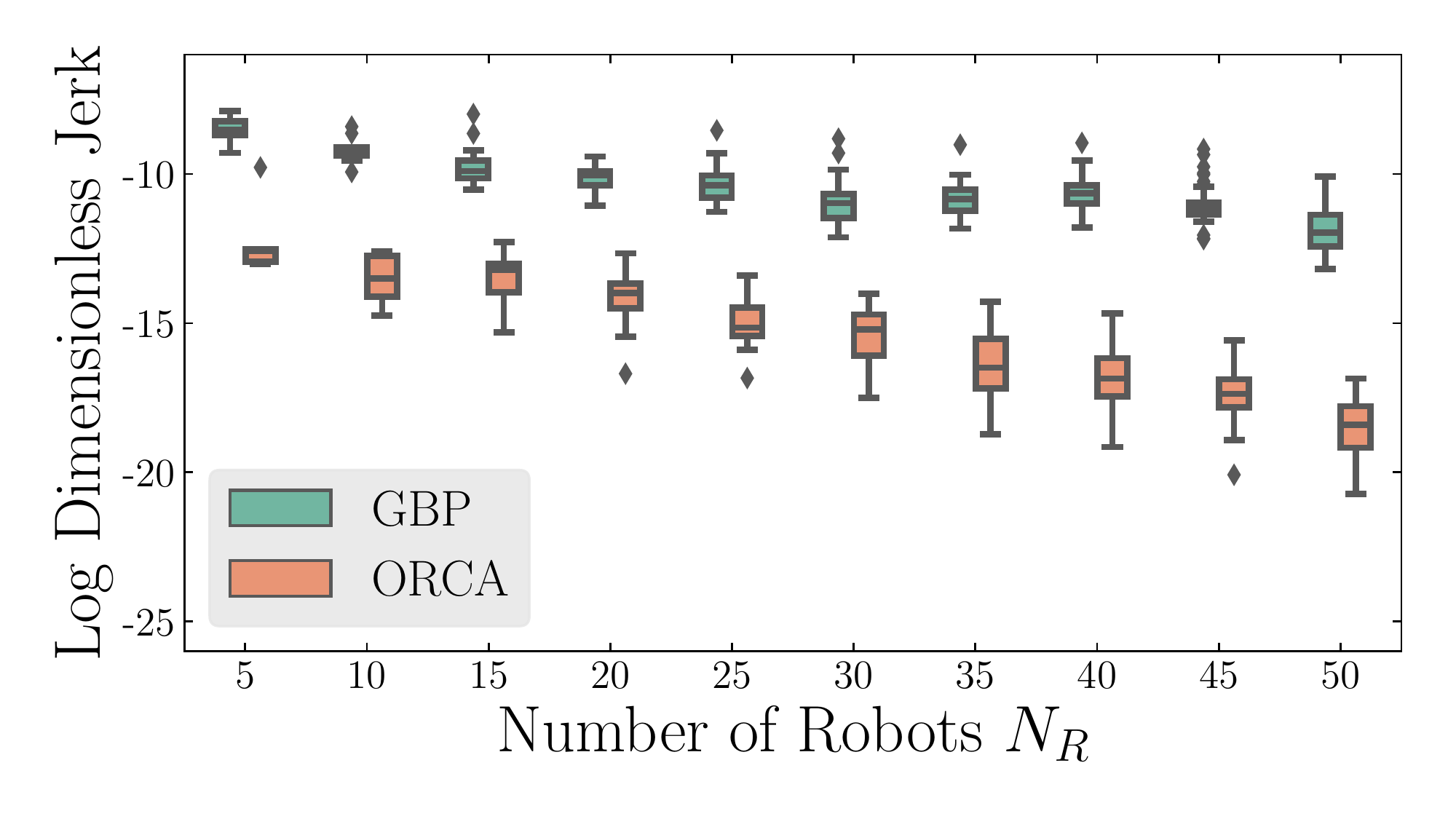}
    \caption{
        Circle experiment: distribution of the LDJ metric as $N_R$ increases, with smoother trajectories shown by more positive values. The worst performing GBP planning robots had smoother paths than the best robots for ORCA.
    }
    \label{fig:LDJ}
\end{figure}

Figure \ref{fig:LDJ} shows the distributions of the LDJ metric at each $N_R$. GBP planning creates trajectories that are far smoother than ORCA. As $N_R$ increases, this difference in performance between our method and ORCA increases. In general the \textit{worst performing} GBP planning robots had trajectories that were smoother than the \textit{best performing} robot using ORCA.

As we are presenting a state planner we assume that robots can perfectly execute their planned trajectories. With real-world robots this can be helped if the planned trajectories are smooth, efficient and induce minimal jerk.

The solid lines in figure \ref{fig:makespan} depict the makespan as $N_R$ is varied. This metric increases in an approximately linear fashion with GBP, and performs better than ORCA for higher $N_R$. ORCA performs better at when $N_R$ is low --- robots reach their goals in shorter amounts of time. However this is at the expense of unrealistic jerky trajectories.

We consider two further experiments in the circle scenario.
\subsubsection {Environment obstacles}
\begin{figure}
    \centering
    \includegraphics[width=0.85\linewidth]{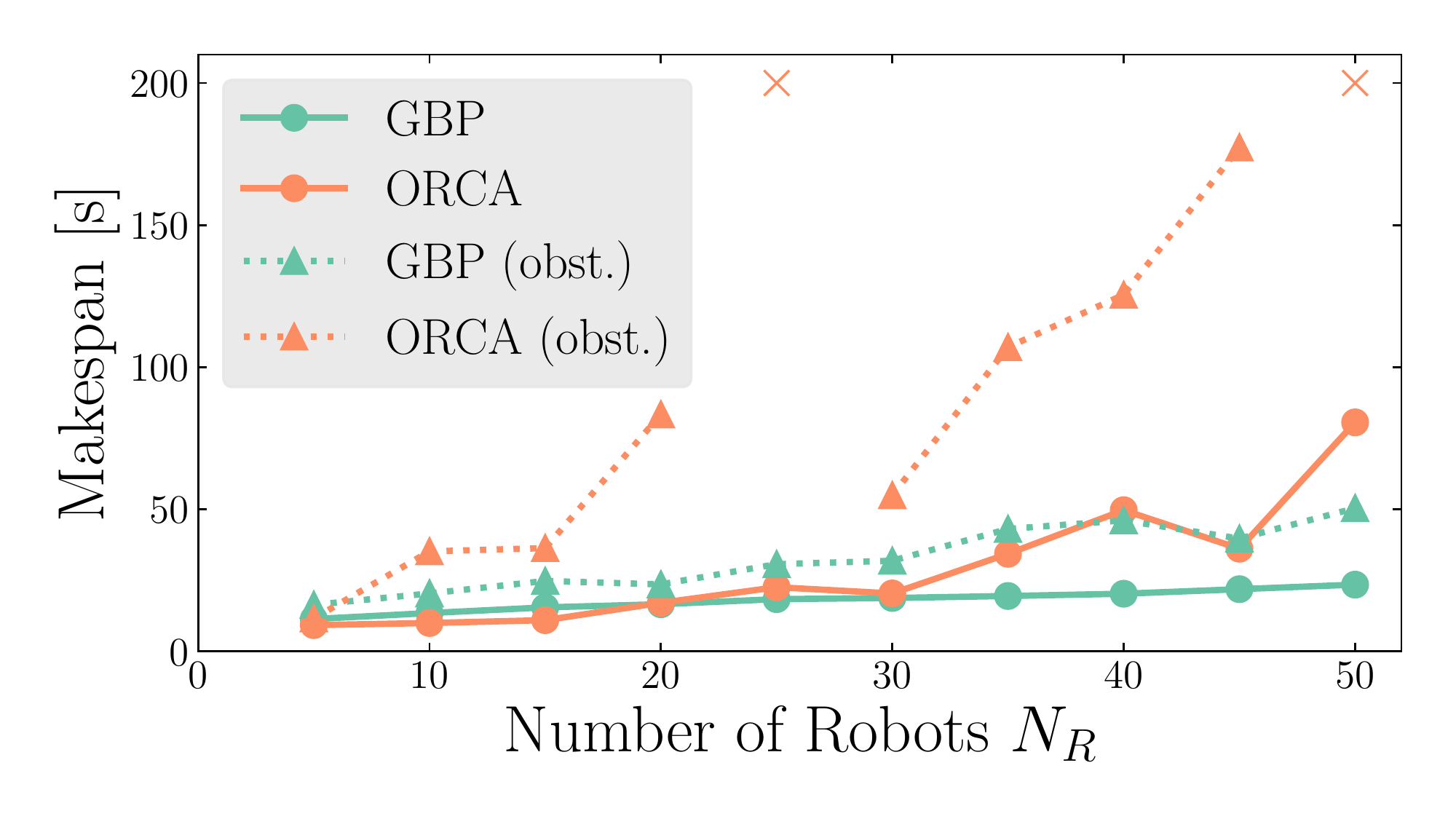}
    \caption{
        Circle experiment with (solid) and without (dotted) obstacles: makespan as $N_R$ increases. Crosses denote trials where not all robots reached their goals.
    }
    \label{fig:makespan}
\end{figure}

Also shown in figure \ref{fig:makespan} with dotted lines are the makespans for ORCA and our GBP planner when 5 polygonal obstacles are placed in the middle of the circle. The paths can be seen in figure \ref{fig:paths_N30}. The results are for one layout of obstacles averaged over 5 seeds. For $N_R=25$ and $30$ some robots using ORCA became deadlocked with the obstacle configuration.
Our method performs well with obstacles, producing makespans that are only slightly higher than in those in free space.

\begin{table}[ht]
  \caption{Effect of varying communication range $r_C$ in the circle experiment with obstacles}
  \label{table:rc_variation}
\centering
\resizebox{\columnwidth}{!}{%
\begin{tabular}{cccc}
\toprule
$r_C$ [m] & Makespan [s] & Mean distance travelled [m] & Log Dimensionless Jerk \\ \midrule
20                                & 12.0       & 104.0      & -9.02 \\
40                                & 12.3      & 104.5        & -8.76 \\
60                                & 12.7      & 104.0        & -8.38 \\
80                                & 14.8      & 103.8       & -8.47 \\
\end{tabular}
}
\end{table}
\subsubsection{Varying network connectivity}
Robots within a communication range $r_C$ of each other form a partially connected network, and can collaboratively modify their planned paths. We investigate the effect of varying $r_C$ for $N_R=30$ for the $100$~m diameter circle formation with obstacles.
Table \ref{table:rc_variation} shows that as $r_C$ increases robots take more of their neighbours into account, resulting in greater makespans but small changes in the distances travelled and path smoothness. This highlights the applicability of our method to real networks where sensing and communication range may be limited.

\subsection{Junction Experiment}
Robots working in crowded environments may have to operate at high speeds with high levels of coordination e.g. traversing junctions between shelves a warehouse.
We simulate one such junction with channel widths of $16$~m and robots moving at $15$~m/s with horizon $t_{K-1}=2$~s. We set $\sigma_d=0.5$~m.

Multirobot systems must be able to maintain a high flowrate of robots without blockages occurring in the junction. We vary the rate $Q_{in}$ at which robots enter the central section of the junction and measure the rate $Q_{out}$ at which they exit and compare our method against ORCA. We choose the central section of the junction to measure flow over 500 timesteps to represent steady-state flow behaviour. 
Robots must exit the junction in the same direction that they entered, without collision; we refer to this as the `correctness condition'.

\begin{figure}
    \centering
    \includegraphics[width=0.85\linewidth]{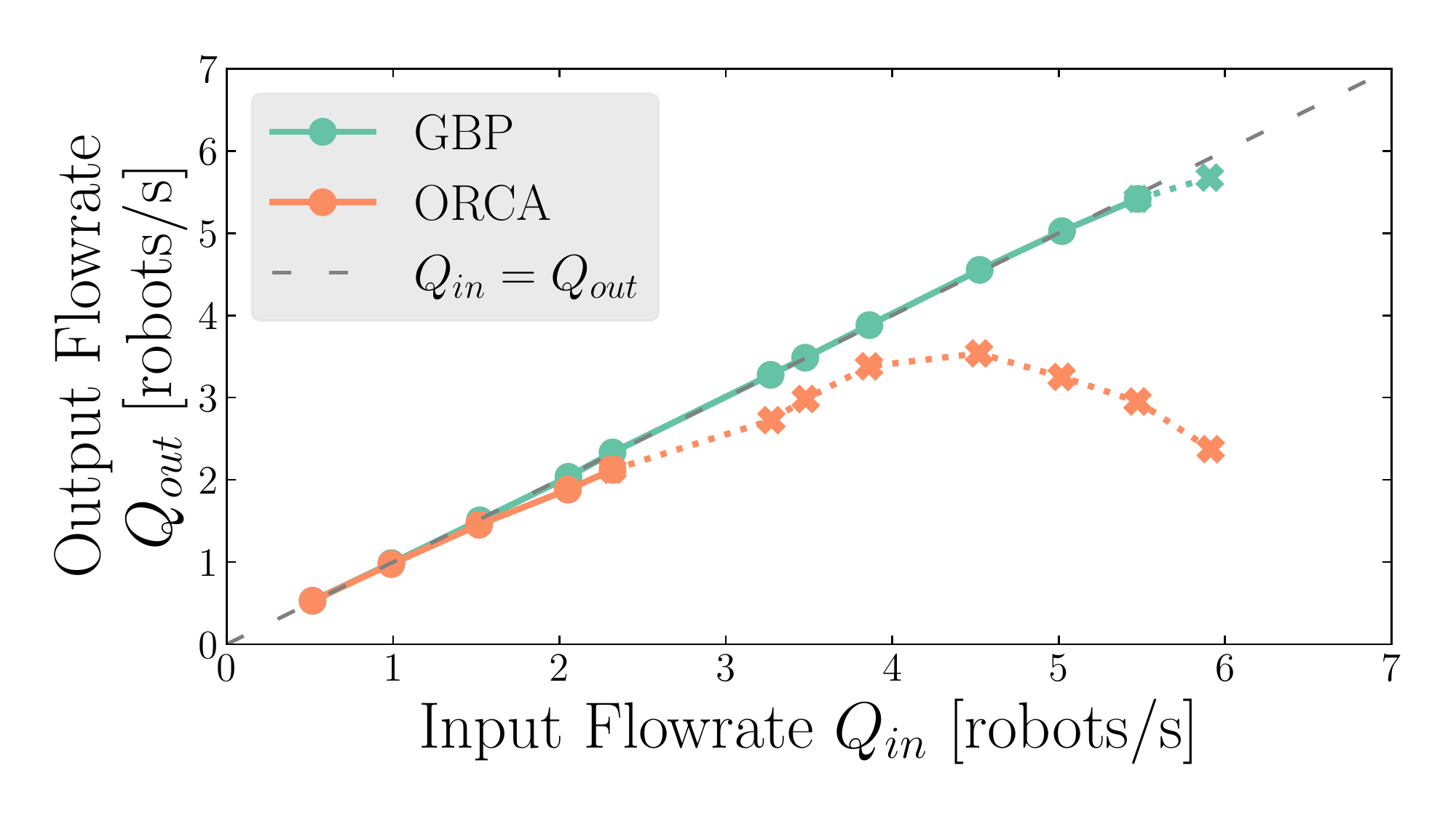}
    \caption{
        Junction experiment: input vs output flowrates. Robots cross the junction at $15$~m/s. Points shown as crosses denoted trials where the correctness condition was violated. GBP planning can sustain flowrates of robots up to $Q_{in}=5.5$~robots/s, compared to ORCA which breaks down at $Q_{in}=2.3$~robots/s.
    }
    \label{fig:flowrates}
\end{figure}

\begin{figure}
    \centering
    \includegraphics[width=0.75\linewidth]{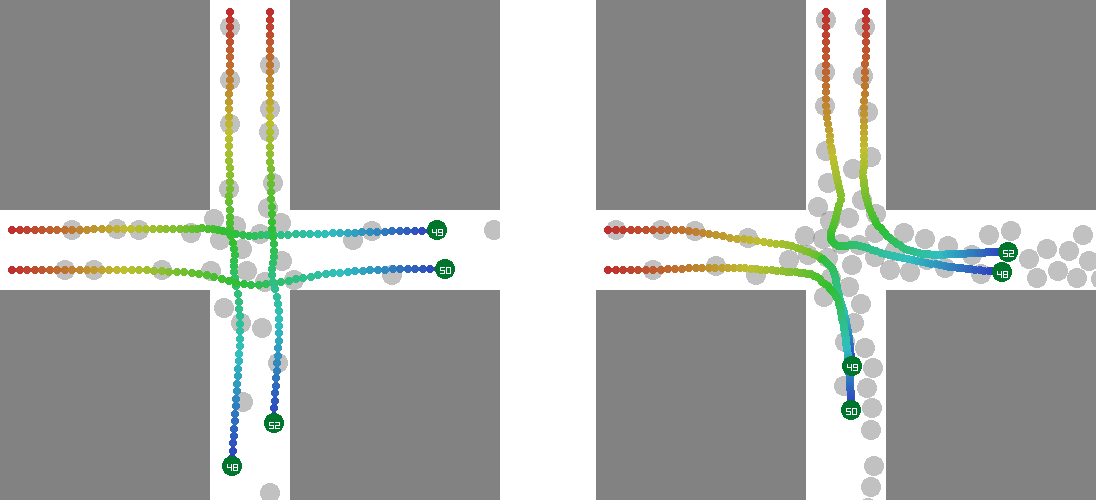}
    \caption{Qualitative comparison of our GBP planner (left) and ORCA (right) in the junction experiment at a high ($Q_{in}=5.5$~robots/s) flowrate. Selected robots are shown in colour. With GBP planning robots can smoothly swerve past each other whereas with ORCA robots block the junction and are deflected.
    }
    \label{fig:junction_paths}
\end{figure}

Figure \ref{fig:flowrates} shows $Q_{out}$ against $Q_{in}$ for GBP planning and ORCA. A dashed line representing the ideal case of equal input and output flowrates ($Q_{out}=Q_{in}$) is also shown for comparison. Trials where the correctness condition was violated are shown as crosses. GBP planning is able to sustain flowrates of up to $Q_{in}=5.5$~robots/s (figure \ref{fig:junction_paths}), compared to ORCA which begins to break down at $Q_{in}=2.3$~robots/s. This is due to GBP planning allowing robots to communicate, and to collaboratively negotiate their planned trajectories with neighbouring robots before they are close to each other.

It is important to note that at high enough flowrates, the effectiveness of both GBP planning and ORCA breaks down  due to the dense packing of incoming robots.

\subsection{Communications Failure Experiment}

Our GBP planner relies on per-timestep peer-to-peer communication between robots. It is assumed that each robot follows a protocol similar to \cite{Murai:etal:ARXIV2022}; it always broadcasts its state information. We consider a communications failure scenario where a robot is not able to receive messages from robots it is connected to. We would expect more cautious behaviour when planning a trajectory.

We simulate a communication failure fraction $\gamma$: at each timestep the robot cannot receive any messages from a randomly sampled proportion $\gamma$ of its connected neighbours. We repeat the circle experiment with 21 robots at two different initial speeds of $10$~m/s and $15$~m/s, measuring the makespan.
The reported result is an average over 5 different random seeds. To be fair, at any timestep for any robot, the failed communications are exactly the same given a fixed seed for both initial velocities considered.

\begin{table}[ht]
\vspace{2mm}
  \caption{A comparison of makespans for the circle experiment with 21 robots for two initial velocities under varying rates of communication failure. Values for mean number of collisions were taken over 5 random seeds.}
  \label{table:comm_failure}
\centering
\resizebox{\columnwidth}{!}{%

\begin{tabular}{ccccc}
\toprule
Initial speed [m/s] & \multicolumn{2}{c}{10} & \multicolumn{2}{c}{15} \\ \midrule
Comm. failure $\gamma$ [\%]                                             & Makespan [s]        & Mean num. collisions       & Makespan [s]        & Mean num. collisions      \\ \midrule
0                                                   & 19.5       & 0.0        & 14.9       & 0.0       \\
10                                                  & 20.3       & 0.0        & 17.1       & 0.0       \\
20                                                  & 22.9       & 0.0        & 18.9       & 0.0       \\
30                                                  & 25.7       & 0.0        & 22.5       & 0.0       \\
40                                                  & 30.8       & 0.0        & 26.5       & 0.0       \\
50                                                  & 35.6       & 0.0        & 30.6       & 0.0       \\
60                                                  & 42.0       & 0.0        & 38.8       & 0.2       \\
70                                                  & 51.3       & 0.0        & 44.6       & 0.8       \\
80                                                  & 87.4       & 0.0        & 63.4       & 0.8       \\
90                                                  & 146.9       & 1.6        & 12.6       & 4.6       \\ \bottomrule
\end{tabular}
}
\end{table}

Table \ref{table:comm_failure} shows that as $\gamma$ increases, it takes longer for all robots to reach their goals. However, trajectories for the $10$~m/s case are completely collision free up to $\gamma=80$\%. As the initial speed increases, collisions happen at lower values of $\gamma$ as robots have less time to react to faster moving neighbours who they may not be receiving messages from.

This experiment shows one of the benefits of GBP --- safe trajectories can still be planned even with possible communication failures, which is likely in any realistic settings. 

\section{CONCLUSIONS}
We have shown that our collaborative method for short-term trajectory planning using Gaussian Belief Propagation (GBP) can ensure smooth, efficient and safe trajectories in problems where a high degree of coordination is required. By performing message passing between robots in a purely peer-to-peer manner, there is no need for a centralised solver; the distributed nature of the problem could be scaled up to large numbers of robots.

Our current approach, like most other planners, assumes that all robots have perfect knowledge of their localisation.
However it is clear that our planning approach could be combined with the multi-robot localisation method based on GBP in \cite{Murai:etal:ARXIV2022} which uses the same communication and computation pattern, to enable true infrastructure-free multi-robot operation.

In the future, we will explore other applications of GBP trajectory planning, such as the coordination of larger numbers of robots which can organise themselves into useful formations.

\addtolength{\textheight}{-12cm}   




\section*{ACKNOWLEDGEMENTS}

Research presented in this paper has been supported by Dyson Technology Ltd. We are grateful for discussions with members of  the Dyson Robotics Lab and Robot Vision Group at Imperial College; particularly Joseph Ortiz, Tristan Laidlow and Ignacio Alzugaray.


\bibliographystyle{unsrt} 
\bibliography{bib,robotvision}

\end{document}

%% file: symbols.tex
\newcommand{\ignore}[1]{}

\newcommand{\ba}{\begin{array}}
\newcommand{\ea}{\end{array}}
\newcommand{\bc}{\begin{center}}
\newcommand{\ec}{\end{center}}
\newcommand{\be}{\begin{enumerate}}
\newcommand{\ee}{\end{enumerate}}
\newcommand{\bea}{\begin{eqnarray}}
\newcommand{\eea}{\end{eqnarray}}
\newcommand{\beas}{\begin{eqnarray*}}
\newcommand{\eeas}{\end{eqnarray*}}
\newcommand{\beq}{\begin{equation}}
\newcommand{\eeq}{\end{equation}}
\newcommand{\bfig}{\begin{figure}}
\newcommand{\efig}{\end{figure}}
\newcommand{\bi}{\begin{itemize}}
\newcommand{\ei}{\end{itemize}}
\newcommand{\bpic}{\begin{picture}}
\newcommand{\epic}{\end{picture}}
\newcommand{\btabular}{\begin{tabular}}
\newcommand{\etabular}{\end{tabular}}
\newcommand{\btable}{\begin{table}}
\newcommand{\etable}{\end{table}}

\newcommand{\es}{\vfill
                 \rule[-6mm]{170mm}{0.7mm} \\
                 \redw{{\tiny
		  \hfill S-\theslide}}
                 \end{slide}}

\newcommand{\vecXX}[1]{{\mathbf {#1}}}

\providecommand{\etal}{{\em et~al.}}

\def \hbar {{\bar{h}}}

\def \vecm {{\vecXX{m}}}

\makeatletter
\renewcommand*\env@matrix[1][*\c@MaxMatrixCols c]{%
  \hskip -\arraycolsep
  \let\@ifnextchar\new@ifnextchar
  \array{#1}}
\makeatother

%% file: output.bbl
\begin{thebibliography}{10}

\bibitem{Soria:etal:NMI2021}
E.~Soria, F.~Schiano, and D.~Floreano.
\newblock Predictive control of aerial swarms in cluttered environments.
\newblock {\em Nature Machine Intelligence}, 3:545--554, 2021.

\bibitem{vandeberg_priority}
J.~van~den Berg and M.H. Overmars.
\newblock Prioritized motion planning for multiple robots.
\newblock In {\em International Conference on Intelligent Robots and Systems
  (IROS)}, 2005.

\bibitem{ORCA}
J.~van~den Berg, S.~Guy, M.~Lin, and D.~Manocha.
\newblock Reciprocal {N}-body collision avoidance.
\newblock {\em {Proceedings of the International Symposium on Robotics Research
  (ISRR)}}, 2009.

\bibitem{AlonsoMora:etal:TRO2018}
J.~Alonso-Mora, P.~Beardsley, and R.~Siegwart.
\newblock Cooperative collision avoidance for noholonomic robots.
\newblock {\em {{IEEE} Transactions on Robotics ({T-RO})}}, 2(2):404--420,
  2018.

\bibitem{Senbaslar:etal:DARS2019}
B.~Senbaslar, W.~Honig, and N.~Ayanian.
\newblock Robust trajectory execution for multi-robot teams using distributed
  real-time replanning.
\newblock In {\em Proceedings of Distributed and Autonomous Robotics Systems},
  2019.

\bibitem{vanparys}
R.~Van~Parys and G.~Pipeleers.
\newblock Online distributed motion planning for multi-vehicle systems.
\newblock In {\em Proceedings of the European Control Conference (ECC)}, 2016.

\bibitem{Luis:etal:RAL2020}
C.~Luis, M.~Vukosavljev, and A.~Schoellig.
\newblock Online trajectory generation with distributed model predictive
  control for multi-robot motion planning.
\newblock {\em IEEE Robotics and Automation Letters}, PP:1--1, 01 2020.

\bibitem{prorok}
Q.~Li, F.~Gama, A.~Ribeiro, and A.~Prorok.
\newblock Graph neural networks for decentralized multi-robot path planning.
\newblock In {\em International Conference on Intelligent Robots and Systems
  (IROS)}, 2020.

\bibitem{Mukadam:etal:IJRR2018}
M.~Mukadam, J.~Dong, X.~Yan, F.~Dellaert, and B.~Boots.
\newblock Continuous-time gaussian process motion planning via probabilistic
  inference.
\newblock {\em {International Journal of Robotics Research ({IJRR})}}, 37(11),
  2018.

\bibitem{Dellaert:AR2021}
F.~Dellaert.
\newblock Factor graphs: Exploiting structure in robotics.
\newblock {\em Annual Review of Control, Robotics, and Autonomous Systems},
  4:141--166, 2021.

\bibitem{Murai:etal:ARXIV2022}
R.~Murai, J.~Ortiz, S.~Saeedi, P.H.J. Kelly, and A.~J. Davison.
\newblock A robot web for distributed many-device localisation.
\newblock {\em arXiv preprint arXiv:2202.03314}, 2022.

\bibitem{Ortiz:etal:ARXIV2021}
J.~Ortiz, T.~Evans, and A.~J. Davison.
\newblock A visual introduction to {Gaussian Belief Propagation}.
\newblock {\em arXiv preprint arXiv:2107.02308}, 2021.

\bibitem{Davison:Ortiz:ARXIV2019}
A.~J. Davison and J.~Ortiz.
\newblock {FutureMapping 2: Gaussian Belief Propagation for Spatial AI}.
\newblock {\em arXiv preprint arXiv:1910.14139}, 2019.

\bibitem{LDJ_paper}
S.~Balasubramanian, A.~Melendez-Calderon, A.~Roby-Brami, and E.~Burdet.
\newblock On the analysis of movement smoothness.
\newblock {\em {Journal of NeuroEngineering and Rehabilitation}}, 12(112),
  2015.

\end{thebibliography}
